\documentclass[preprint,12pt,sort&compress]{elsarticle}

\usepackage{amssymb,subcaption,amsmath,float,paralist}

\usepackage[linesnumbered,ruled,vlined]{algorithm2e}

\usepackage{xcolor,soul}

\SetCommentSty{mycommfont}
\usepackage{arydshln}
\usepackage{paralist}
\journal{}
\date{}
\usepackage[margin=1.0in]{geometry}
\usepackage{wrapfig}
\usepackage{listings}

\definecolor{codegreen}{rgb}{0,0.6,0}
\definecolor{codegray}{rgb}{0.5,0.5,0.5}
\definecolor{codepurple}{rgb}{0.58,0,0.82}
\definecolor{backcolour}{rgb}{0.95,0.95,0.92}
\lstdefinestyle{mystyle}
{
  backgroundcolor=\color{backcolour},   
  commentstyle=\color{codegreen},
  keywordstyle=\color{magenta},
  numberstyle=\tiny\color{codegray},
  stringstyle=\color{codepurple},
  basicstyle=\sffamily\small,
  breakatwhitespace=false,         
  breaklines=true,                 
  captionpos=b,                    
  keepspaces=true,                 
  numbers=left,                    
  numbersep=5pt,                  
  showspaces=false,                
  showstringspaces=false,
  showtabs=false,                  
  tabsize=2
}
\begin{document}

\begin{frontmatter}

% \title{Estimating injection rate from ground displacement in coupled flow and geomechanics problems using deep learning enabled Bayesian inference}

\title{Comparison of LSTM autoencoder based deep learning enabled Bayesian inference using two time series reconstruction approaches}

\author[inst1]{Saumik Dana}
%\author[inst1]{Birendra Jha}
%\author[inst2]{Karthik Lyathakula}
\affiliation[inst1]{organization={University of Southern California}}
% \affiliation[inst2]{organization={North Carolina State University}}
            
\begin{abstract}
In this work, we use a combination of Bayesian inference, Markov chain Monte Carlo and deep learning in the form of LSTM autoencoders to build and test a framework to provide robust estimates of injection rate from ground surface data in coupled flow and geomechanics problems. We use LSTM autoencoders to reconstruct the displacement time series for grid points on the top surface of a faulting due to water injection problem. We then deploy this LSTM autoencoder based model instead of the high fidelity model in the Bayesian inference framework to estimate injection rate from displacement input.
\end{abstract}

% \begin{highlights}
% \item A combination of Bayesian inference, Markov chain Monte Carlo and deep learning has been used
% \item Deep learning is used to construct a reduced order model
% \item Bayesian inference with Markov chain Monte Carlo sampling uses this reduced order model
% \item Initial guess for inference is way off the ground truth to allow checking robustness of inference
% \end{highlights}

\begin{keyword}
LSTM autoencoder \sep Bayesian inference \sep Markov chain Monte Carlo \sep coupled flow and geomechanics
\end{keyword}

\end{frontmatter}

\section{Introduction}

The high fidelity forward model for coupled flow and geomechanics~\cite{danacg,danacmame,danajcp,danalanl,danaperformance,danasiam,danathesis,GASPARINI2021101330} in the absence of inertia is a piece of the puzzle in which forward simulations are used to arrive at the seismic impact of fault slip on earthquake activity. Typically, the earthquakes are measured with seismograms/geophones on the surface as P-waves and S-waves, and then these readings are used to calibrate the seismic activity for constant monitoring. The acceleration/displacement field around the fault slip activity translates to these waves recorded on the surface, and forward simulations with wave propagation bridge that gap. That being said, a seismic recording on the surface cannot easily be backtraced to the source of the fault slip. That is precisely inverse modeling, and the Bayesian framework coupled with Markov chain Monte Carlo~\cite{OLIVIER2020101204} allows us to put such a framework in place. The accelerations/displacements are field quantities, and the inverse estimation of the field around the fault from the time series at the seismogram/geophone is not a trivial task. With that in mind, we test the robustness of the Bayesian/MCMC framework to inversely estimate a value instead of a field. To run Bayesian though, we need multiple (sometimes millions of) high fidelity simulation runs, which will become infeasible if the model is sophisticated. Hence, reduced order models are needed, and we deploy LSTM autoencoders for that purpose. In this work, we go from the forward model to the simulations to the time series data for the ground surface data, then LSTM autoencoder based reconstruction for an archetypal faulting due to injection problem, and then eventually Bayesian inference estimation of injection rate using the constructed reduced order model. The big picture scenario is that the recording at the seismogram/geophone would be fed into the Bayesian/MCMC framework as an input, and the framework would provide an estimate of whichever model parameters are critical. 
% This document is structured as follows: In Section 2, we give details of the forward model and the specific problem under consideration, in Section 3, we give details of the LSTM autoencoder based time series reconstruction, Section 4 provides a formalism of the Bayesian inference framework with Markov chain Monte Carlo, in Section 5 we present the estimate results and in Section 6, we provide conclusions and outlook. 
The procedural framework is elucidated in the following steps
\begin{compactitem}
    \item Run high fidelity simulations for a bunch of injection rates
    \item For each injection rate, construct a displacement time series at a chosen grid point on the ground surface
    \item
Add noise to the time series to eventually serve as noisy data for the Bayesian inference framework
\item
Train the LSTM autoencoder with time stamp and injection rate as input and displacement time series (without the noise) as the target
\item
Run the Bayesian inference framework on the noisy data with the LSTM autoencoder based reduced order model
\item
Test the robustness of the framework by comparing the estimates of the injection rate with the ground truth injection rate
\end{compactitem}

\subsection{Water injection in the presence of fault}

\begin{figure}[htb!]
\begin{subfigure}{.55\textwidth}
\centering
\includegraphics[trim={1.75cm 0 0 5cm},clip,scale=0.425]{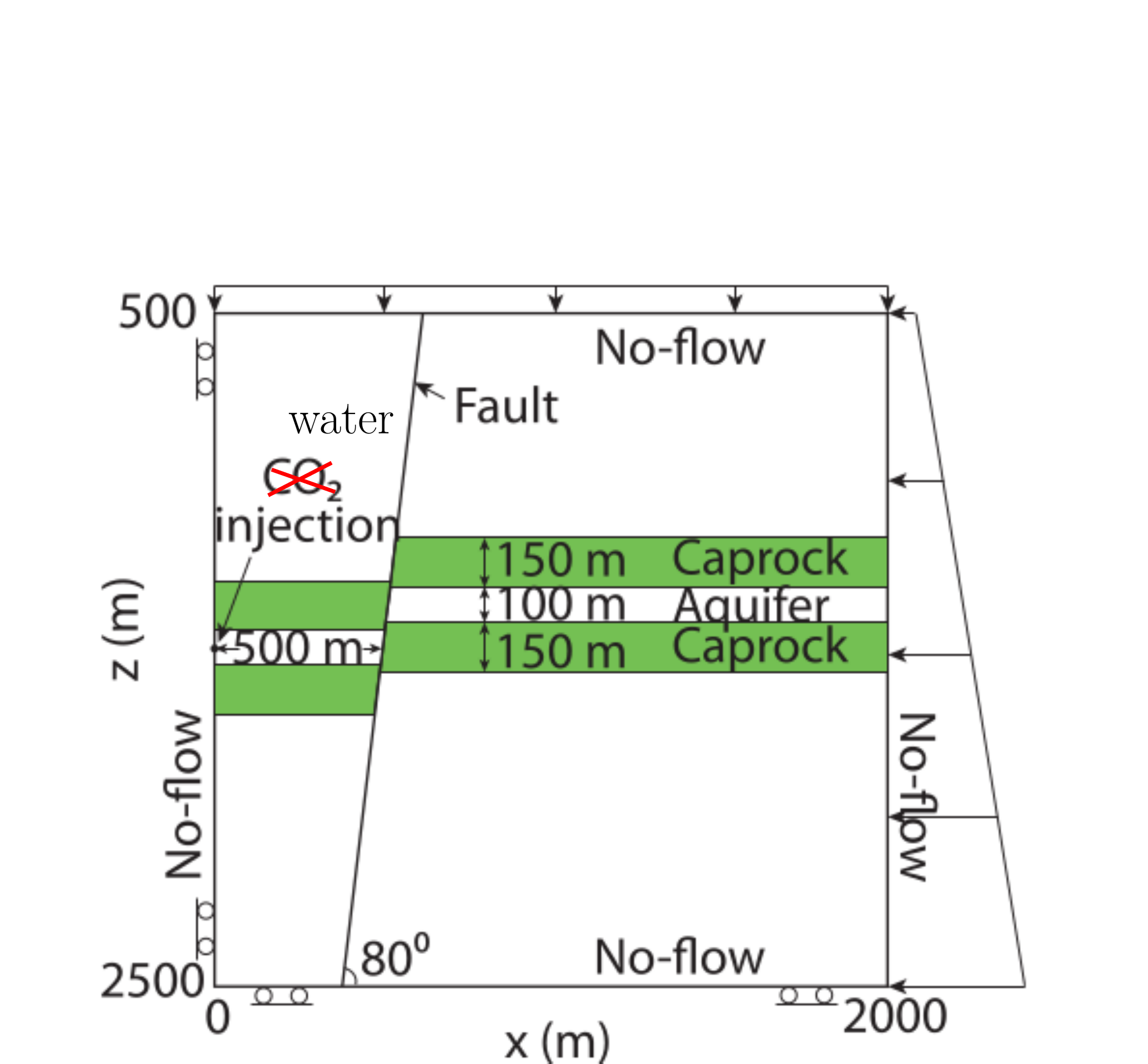}
\end{subfigure}
\begin{subfigure}{.45\textwidth}
\centering
\includegraphics[trim={9cm 2.5cm 0 2.5cm},clip,scale=0.425]{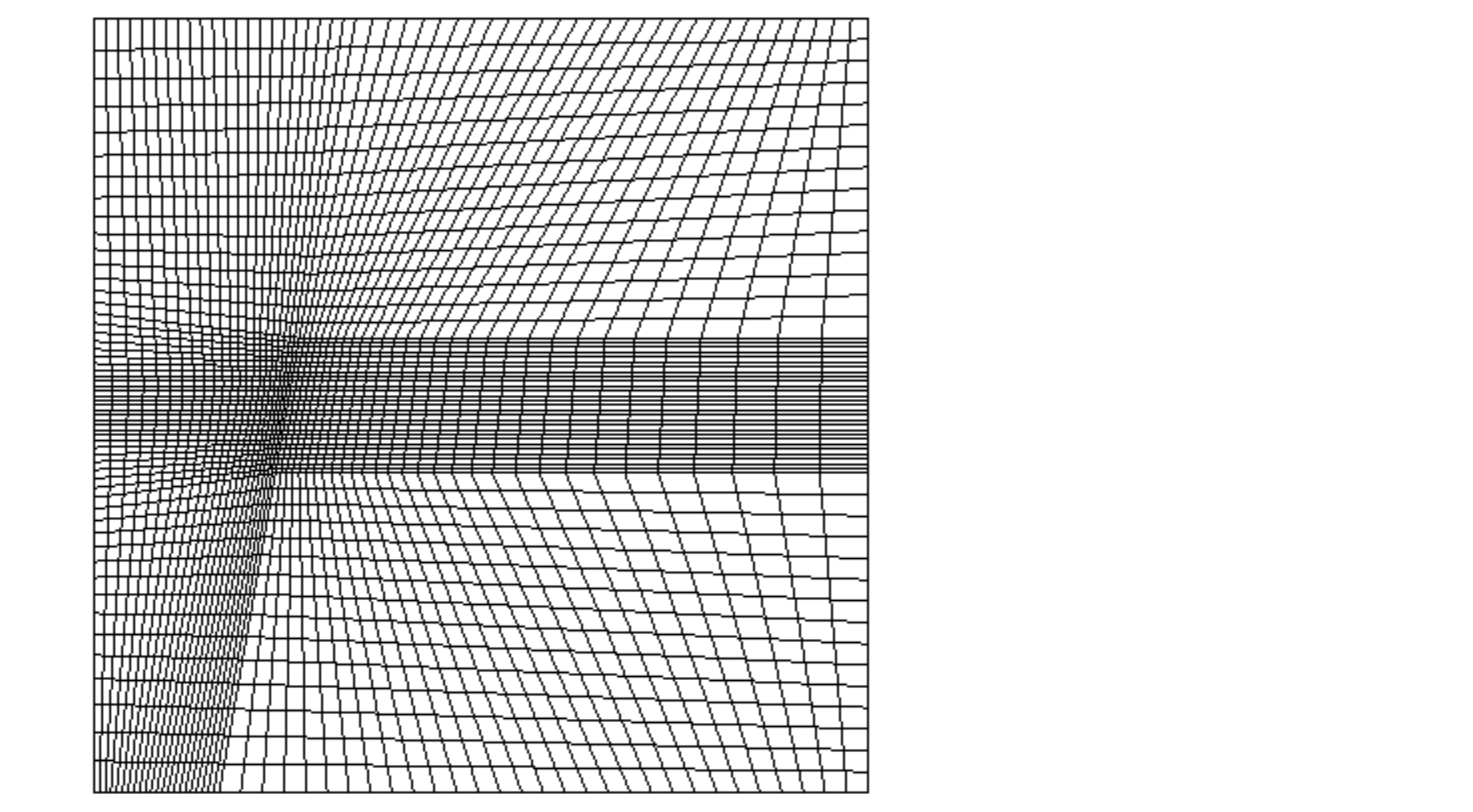}
\end{subfigure}
 \caption{Model of the water injection plane strain case~\cite{CapF2011a}) and the mesh of $3127$ nodes and $3016$ elements with more refinement around the fault}
 \label{f:cappa_geometry}
\end{figure}

\begin{figure}[htb!]
\begin{subfigure}{.5\textwidth}
    \centering
    \includegraphics[trim={7cm 0 9cm 0},clip,scale=0.35]{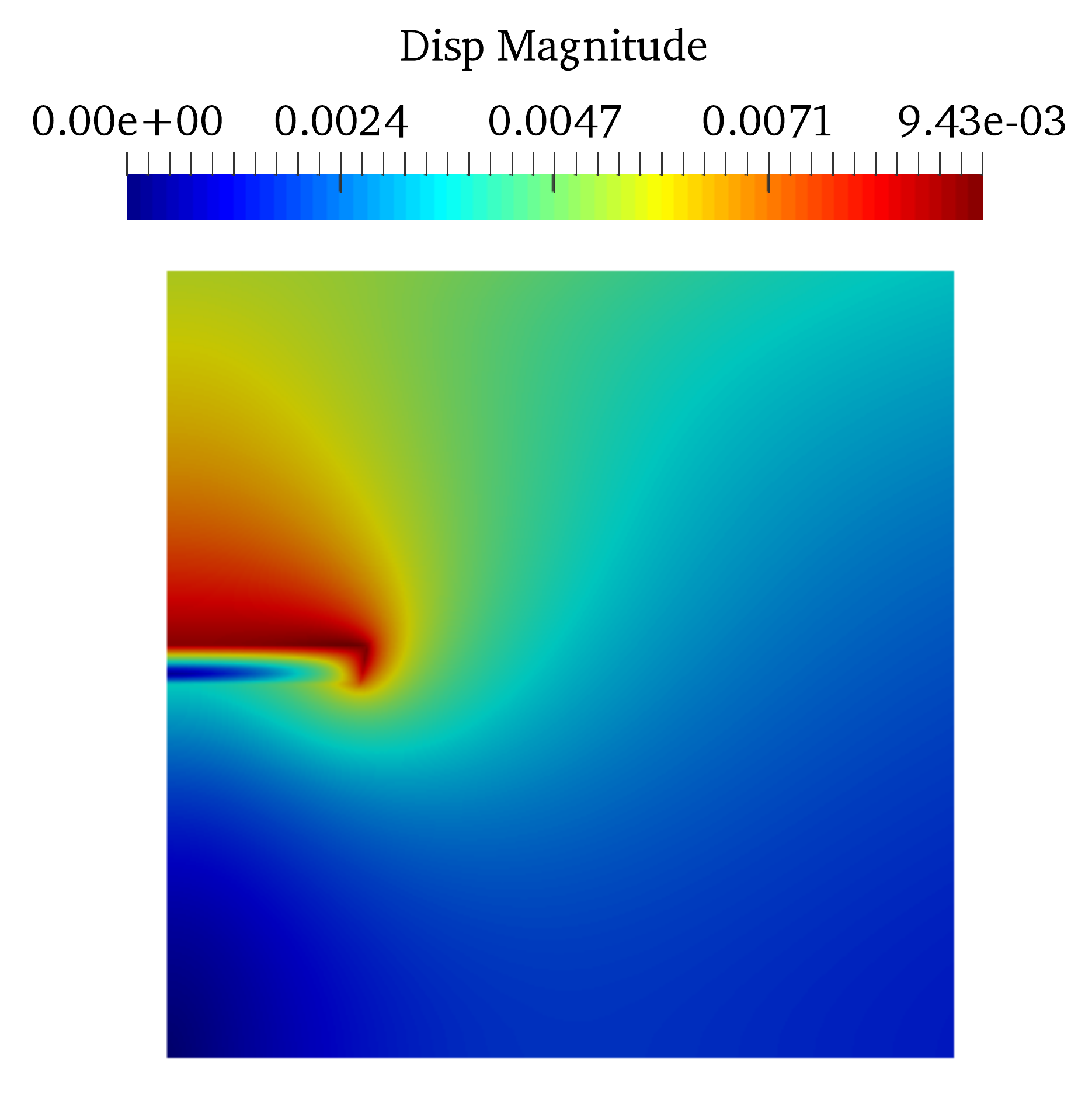}
\end{subfigure}
\begin{subfigure}{.5\textwidth}
    \centering
    \includegraphics[trim={7cm 0 9cm 0},clip,scale=0.35]{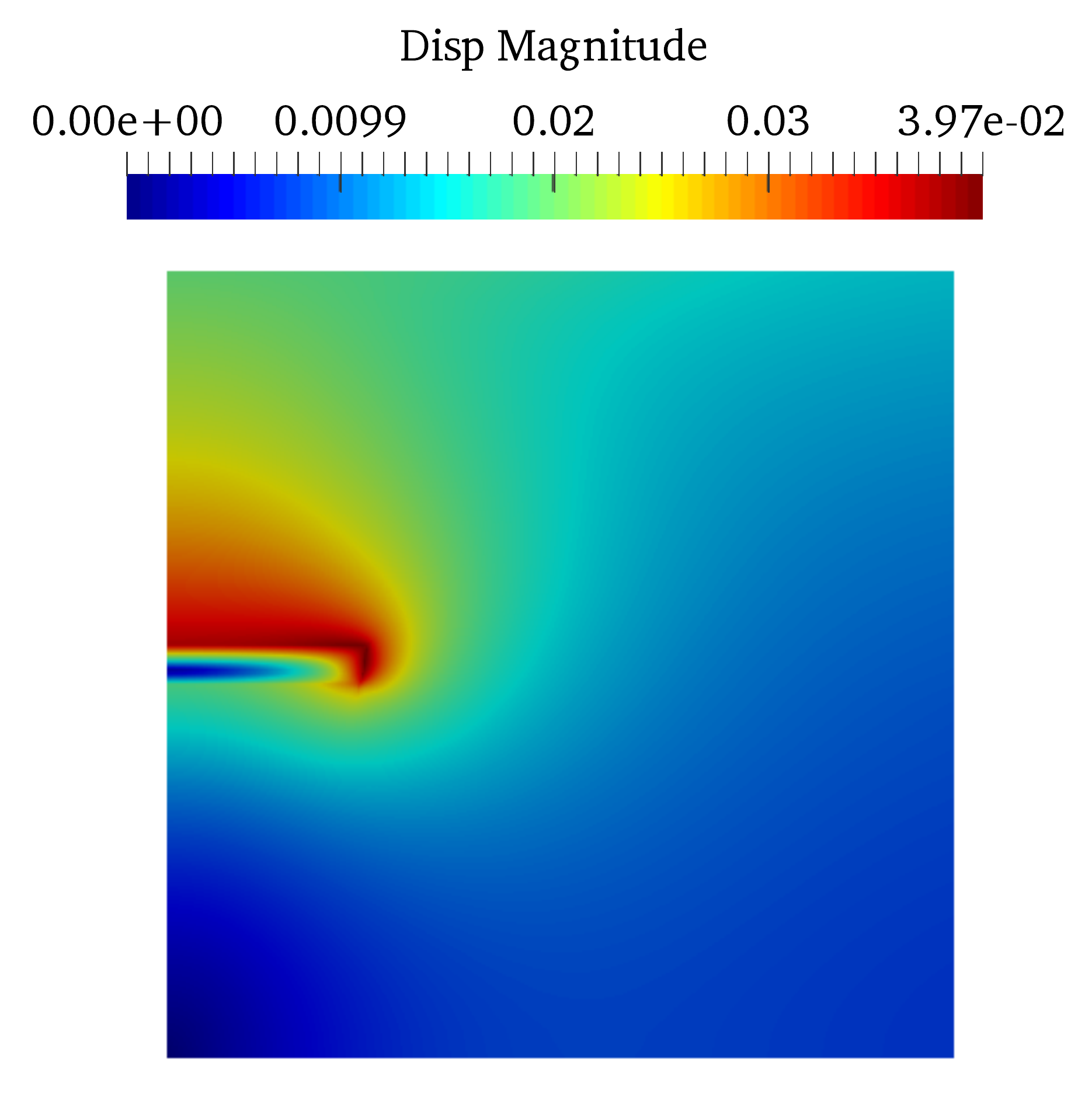}
\end{subfigure}
    \caption{Snapshots of displacement at t=60 days for injection rates of 100 MSCF/day and 400 MSCF/day respectively}
    \label{snapshots}
\end{figure}

\begin{figure}[htb!]
\begin{subfigure}{.5\textwidth}
    \centering
    \includegraphics[scale=0.50]{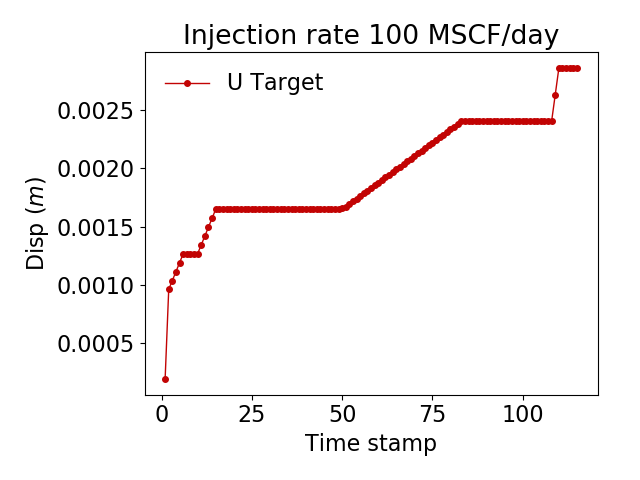}
\end{subfigure}
\begin{subfigure}{.5\textwidth}
    \centering
    \includegraphics[scale=0.50]{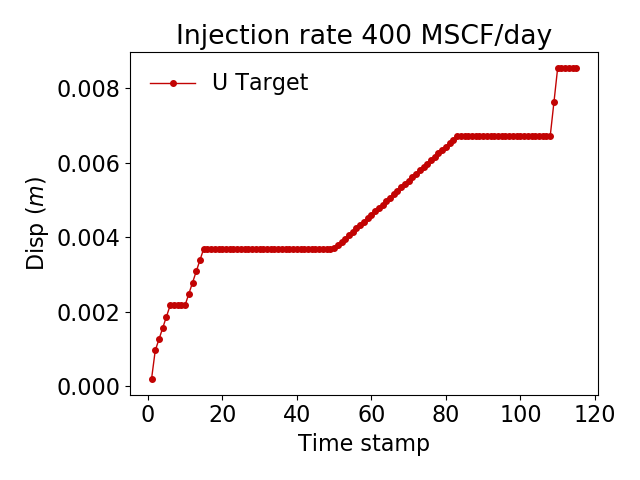}
\end{subfigure}
    \caption{Displacement time series ground truth for $u$ at $(0,500)$ for injection rates on either end of the data spectrum.}
    \label{ground truth}
\end{figure}

As shown in Fig.~\ref{f:cappa_geometry}, we consider a two-dimensional plane-strain model with the fault under normal faulting conditions, that is, the vertical principal stress due to gravity is the largest among the three principal stresses. The mathematics and numerics of the forward model is explained in~\ref{fmdetails}. The aquifer is hydraulically compartmentalized with a sealing fault that cuts across it. The storage capacity of the aquifer is limited by overpressurization and slip on the fault. The initial fluid pressure at $500\,m$ depth is $5\,MPa$ and $24.63\,MPa$ at $2500\,m$, considering a hydrostatic gradient of $9.81\,MPa/km$ and an atmospheric pressure of $0.1\,MPa$ at the ground surface. The rock density is $2260\,kg/m^3$, so the lithostatic gradient is $22.17\, MPa/km$. Assuming a porosity of $0.1$, the initial vertical stress is $11.085\times 0.9+5\times 0.1=10.4765\,MPa$ at $500\,m$, and $22.17
\times 2.5\times 0.9+24.63\times 0.1=52.3455\,MPa$ at $2500\,m$. We choose a value of $0.7$ for the ratio of horizontal to vertical initial total stress. The average bulk density is $\rho_b = 2260\times 0.9+1000\times 0.1=2134 kg/m^3$ , the average sonic compressional and shear velocities are $V_p = 730$ m/s and $V_s = 420$ m/s respectively. The Biot coefficient is assumed to be $b = 1.0$. The friction coefficient drops linearly from  static friction $\mu_s=0.5$ to  dynamic friction  $\mu_d=0.2$ over $d_c=5\,mm$. Snapshots of the displacement evolution are given in Fig.~\ref{snapshots}. The simulator spits out vtk files for each time stamp in the simulation. Our job is to extract the displacement values corresponding to grid points on the top surface, and construct a time series as the ground truth, as shown in Fig.~\ref{ground truth}. A code snippet for processing vtk files is provided in Listing~\ref{pycode1}.

\begin{lstlisting}[language=Python,caption=Code for processing vtk files,label=pycode1]
class parse_vtk:
# Base class for parsing vtk files

   def get_surface_information(self, vector_name):
        """
        :vector_name: the vector you want to process
        :return: displacement components
        """
        reader = vtk.vtkDataSetReader()
        reader.SetFileName(self.infile)
        reader.Update()
        data = reader.GetOutput()
        npoints = data.GetNumberOfPoints()
        d = data.GetPointData()
        array = d.GetArray(vector_name)
        
        u, v, w, x, y, z = np.zeros(npoints),np.zeros(npoints),np.zeros(npoints),np.zeros(npoints),np.zeros(npoints),np.zeros(npoints)
        
        for n in range(npoints):
            x[n], y[n], z[n] = data.GetPoint(n)
            u[n], v[n], w[n] = array.GetTuple(n)
        
        # Surface information at min x and max y
        u = u[np.where((x==min(x)) & (y==max(y)))[0]]
        v = v[np.where((x==min(x)) & (y==max(y)))[0]]
        
        del x, y, z
        return np.sqrt(u**2+v**2)
\end{lstlisting}

\section{Reconstruction using LSTM autoencoders}

A reduced order model would effectively mean reconstructing this time series using LSTM autoencoder, and the optimal deep learning parameters to best reconstruct the time series. The deep learning piece is built on the PyTorch framework, and all simulations are run on a basic AMD Ryzen 3 3200U with Radeon Vega Mobile Gfx × 4 processor. A code snippet is provided in Listing~\ref{pycode2}.

\begin{lstlisting}[language=Python,caption=LSTM autoencoder code snippet,label=pycode2]
class lstm_encoder(nn.Module):
# Encodes time-series sequence
    
    def __init__(self, input_size, hidden_size, num_layers):
        super(lstm_encoder, self).__init__()
        self.lstm = nn.LSTM(input_size = input_size, hidden_size= hidden_size, num_layers = num_layers)
        
    def forward(self, x_input): 
    # called internally by PyTorch
        lstm_out, self.hidden = self.lstm(x_input.view(x_input.shape[0], x_input.shape[1], self.input_size))
        return lstm_out, self.hidden     
    
class lstm_decoder(nn.Module):
# Decodes hidden state output by encoder
    
    def __init__(self, input_size, hidden_size, num_layers):
        super(lstm_decoder, self).__init__()
        self.lstm = nn.LSTM(input_size = input_size, hidden_size = hidden_size, num_layers = num_layers)
        self.linear = nn.Linear(hidden_size, input_size)
        
    def forward(self, x_input, encoder_hidden_states): 
    # called internally by PyTorch
        lstm_out, self.hidden = self.lstm(x_input.unsqueeze(0), encoder_hidden_states)
        output = self.linear(lstm_out.squeeze(0))     
        return output, self.hidden
\end{lstlisting}

\subsection{Nonoverlapping window approach}

\begin{figure}[htb!]
\begin{subfigure}{.5\textwidth}
    \centering
    \includegraphics[scale=0.425]{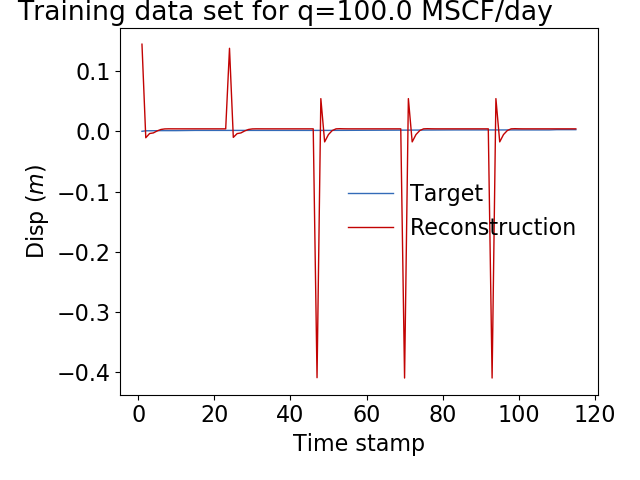}
\end{subfigure}
\begin{subfigure}{.5\textwidth}
    \centering
    \includegraphics[scale=0.425]{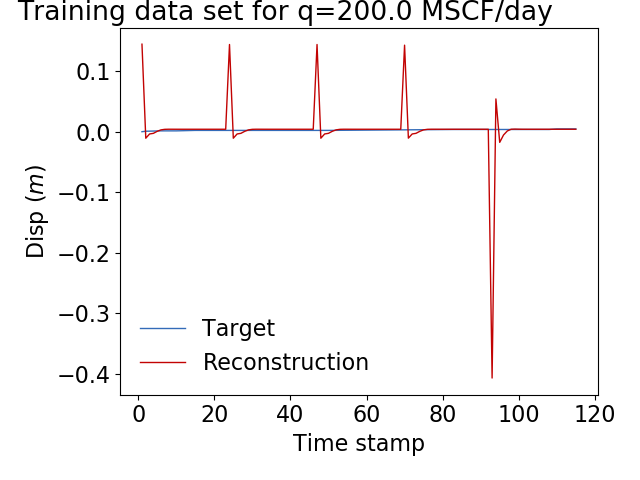}
\end{subfigure}
\begin{subfigure}{.5\textwidth}
    \centering
    \includegraphics[scale=0.425]{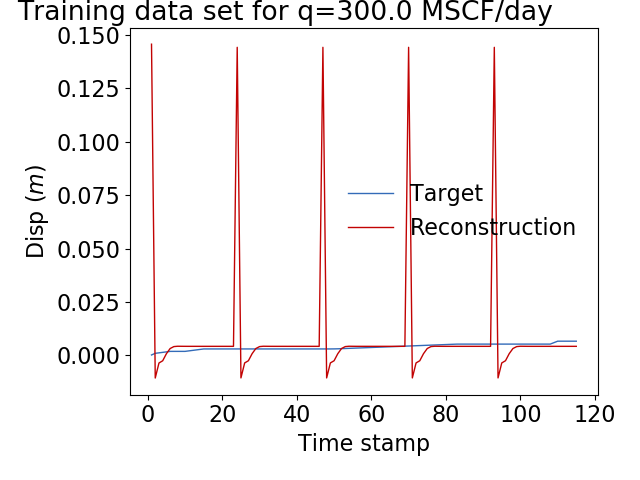}
\end{subfigure}
\begin{subfigure}{.5\textwidth}
    \centering
    \includegraphics[scale=0.425]{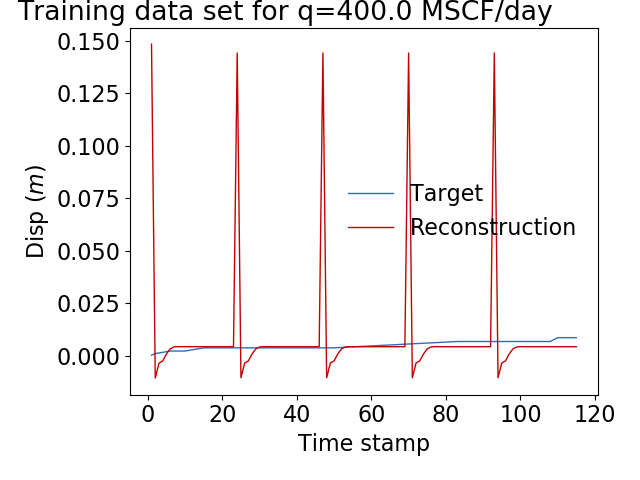}
\end{subfigure}
    \caption{Reconstruction of displacement time series ground truth for $u$ at $(0,500)$ for different injection rates with nonoverlapping window approach}
    \label{reconstruct}
\end{figure}

We use a window size of 23 for 115 time steps, hidden state size of 5, number of LSTM layers per encoder and decoder is 1, and we deploy the Adam optimizer to train the model using only 10 epochs to avoid overfitting. The 115 data points are divided exactly into windows of 23 data points, and the LSTM autoencoder is trained for these windows. During reconstruction, these data chunks in the form of windows are fed into the trained LSTM autoencoder. The ratio of the window to the hidden state size is a measure of the amount of compression that is imposed while encoding the information using the encoder. We observe from Fig.~\ref{reconstruct} that the nonoverlapping window approach causes spikes in the reconstruction at the ends of each window. This is because the reconstruction at the start of a window does not carry information about the time series history from the end of the previous window. In order to smooth these spikes, we present the sliding window approach as explained below

\subsection{Sliding window approach}

\begin{figure}[htb!]
\begin{subfigure}{.5\textwidth}
    \centering
    \includegraphics[scale=0.425]{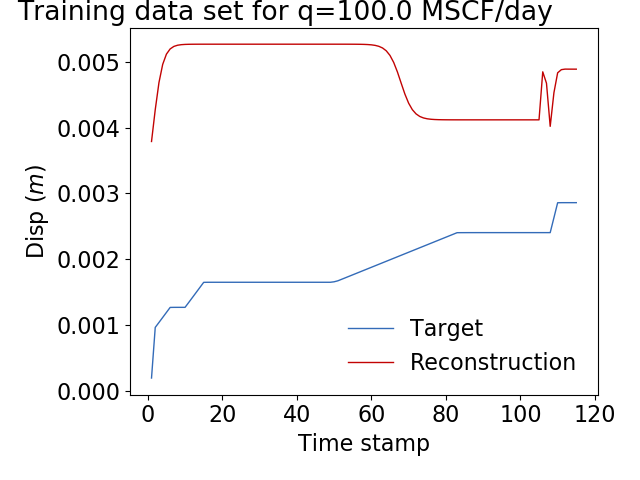}
\end{subfigure}
\begin{subfigure}{.5\textwidth}
    \centering
    \includegraphics[scale=0.425]{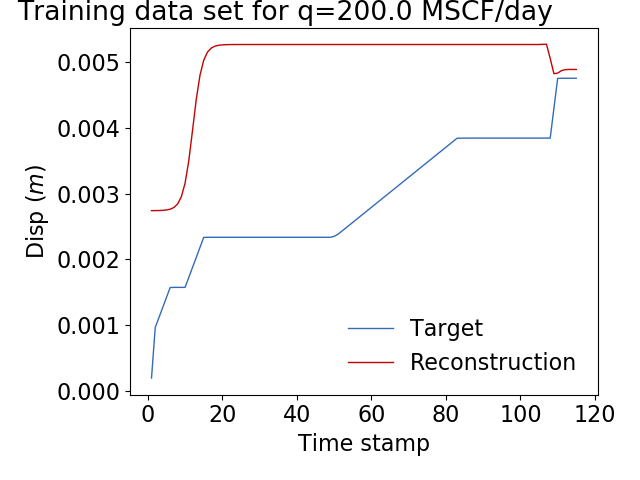}
\end{subfigure}
\begin{subfigure}{.5\textwidth}
    \centering
    \includegraphics[scale=0.425]{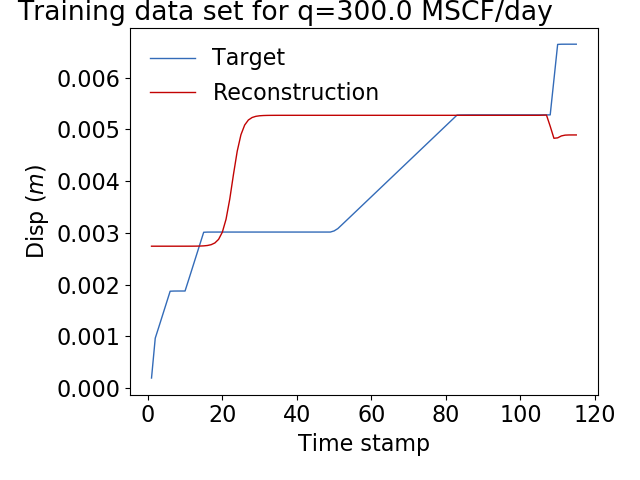}
\end{subfigure}
\begin{subfigure}{.5\textwidth}
    \centering
    \includegraphics[scale=0.425]{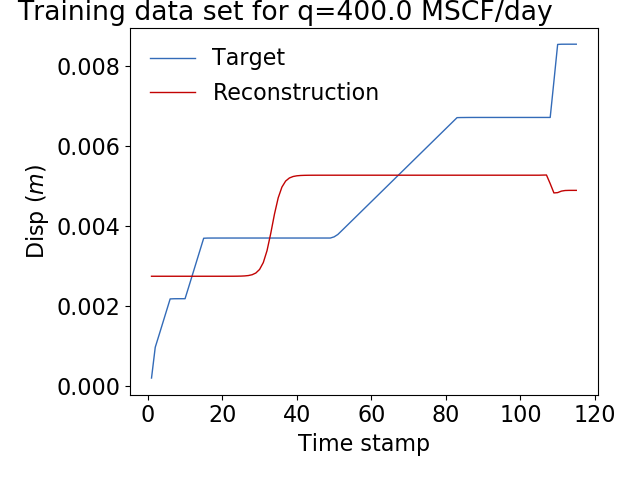}
\end{subfigure}
    \caption{Reconstruction of displacement time series ground truth for $u$ at $(0,500)$ for different injection rates with sliding window approach}
    \label{reconstruct1}
\end{figure}

We use a window size of 10 for 115 time steps, hidden state size of 5, number of LSTM layers per encoder and decoder is 1, and we deploy the Adam optimizer to train the model using only 10 epochs to avoid overfitting. The 115 data points are divided into windows of 10 data points, and each window slides forward by 1, and the LSTM autoencoder is trained for these windows. During reconstruction, these data chunks in the form of windows are fed into the trained LSTM autoencoder. As we observe in Fig.~\ref{reconstruct1}, the sliding window approach allows the reconstructions over these overlaps to be averaged out, which smooths out the spikes that are observed in the nonoverlapping window approach. The reality is that the time series across the injection rates in the spectrum of the generated data spans an order of magnitude, and to fit all that into a LSTM autoencoder in a one size fits all manner is not a trivial task. 

\section{The formalism of Bayesian inference with Markov chain Monte Carlo sampling}

The Bayesian inference framework works on the basic tent of uncovering a distribution centered around the true value and starts off with an initial guess for the distribution also called ``prior'' $\mathcal{D}$ to eventually get to the most accurate distribution possible also called ``posterior'' $\mathcal{P}$ through a likelihood $\mathcal{L}$. The Bayes theorem in a nutshell is:
\begin{align}
\mathcal{P} = \frac{\mathcal{D} \times \mathcal{L}}{\int \mathcal{D} \times \mathcal{L}}
\end{align}
The prior is typically taken to be a Gaussian distribution and the likelihood carries information about the forward model. The quantity that makes evaluation of the posterior difficult is the integral term in the denominator. Since direct evaluation of the integral using quadrature rules is expensive, sampling methods like Markov chain Monte Carlo (MCMC)~\cite{shapiro2003monte,hastings1970monte,haario2001adaptive,mueller2010exploring} are used. To put it mathematically, if we were evaluating an integral, then the sampling would apply to points at which we know the value of integrand, and then proceed to evaluate the integral. But if the integrand at each of those points is a distribution rather than a value, it makes the sampling and subsequent averaging significantly more complicated. By constructing a Markov chain that has the desired distribution as its equilibrium distribution, one can obtain a sample of the desired distribution by recording states from the chain. The more steps are included, the more closely the distribution of the sample matches the actual desired distribution.
\subsection{Applied to our problem}

\begin{figure}[htb!]
\begin{subfigure}{.5\textwidth}
    \centering
    \includegraphics[scale=0.50]{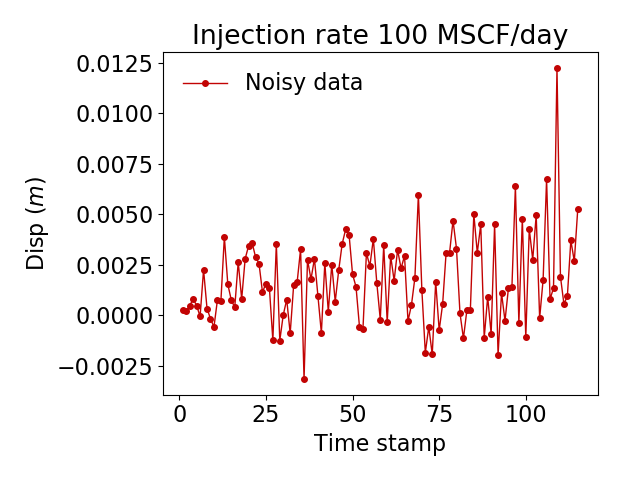}
\end{subfigure}
\begin{subfigure}{.5\textwidth}
    \centering
    \includegraphics[scale=0.50]{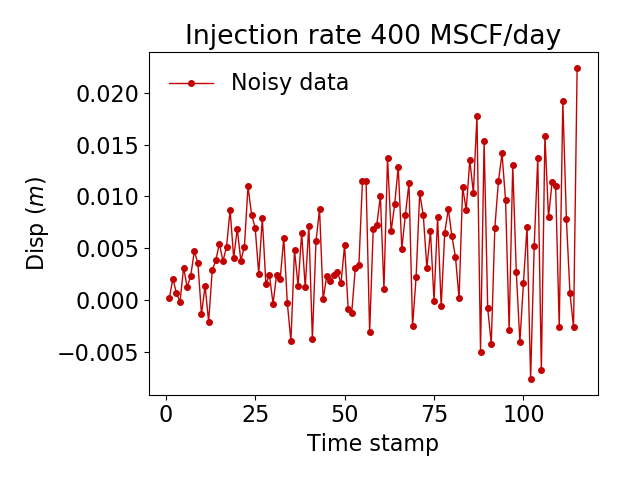}
\end{subfigure}
    \caption{Displacement time series noisy data for $u$ at $(2000,500)$ for injection rates on either end of the data spectrum. The noisy data is generated from the ground truth using a Gaussian distribution}
    \label{noisy}
\end{figure}

In this particular inverse problem, the displacement response of the model is known and the goal is to estimate the injection rate $q$. To formalize the problem, consider the relationship between displacement $u(t)$ and the forward model $\mathcal{F}(\theta)$ with model parameters, constants and variables $\theta$ by the following statistical model
\begin{align}
\label{e:ratestate_stat}
u(t) = \mathcal{F}(\theta)+\epsilon
\end{align}
where $\epsilon$ is the noise. Assuming the $\epsilon \sim N(0,\sigma^2)$ as unbiased, independent and identical normal distribution with standard deviation $\sigma$ allows us to conveniently generate the synthetic data as shown in Fig.~\ref{noisy}. The goal of the inverse problem is to determine the model parameter distribution as follows
\begin{align}
\label{e:ratestate_bayes}
\pi(q\vert u(t_1),...,u(t_n))= \frac{\pi(u(t_1),...,u(t_n)\vert q) \pi_0(q)}{\int_{q} \pi(u(t_1),...,u(t_n)\vert q) \pi_0(q) dq}
\end{align}
where $\pi_0(q)$ is the prior distribution and $\pi(u(t_1),...,u(t_n)\vert q)$ is the likelihood given by
\begin{align}
\label{e:ratestate_likeli}
&\pi(u(t_1),...,u(t_n)\vert q) =  \prod_{i=1}^{n} \pi(u(t_i)\vert q) = \prod_{i=1}^{n} \frac{1}{\sigma \sqrt{2\pi}} e^{-\frac{1}{2} \left(\frac{u(t_i)-\mathcal{F}(\theta)}{\sigma}\right)^2}
\end{align}

\subsection{Adaptive metropolis algorithm}

\begin{algorithm}[htb!]
\caption{Metropolis-Hastings algorithm}
\label{mh}
m $\gets 0$\;
$q^m \gets q^0$\tcp*{$q^0$ is an initial guess}
$\boldsymbol{V}\sim q^m$\tcp*{Construct covariance matrix}
\For{j=1,2,...,n}{
$q^*\sim q^m,\boldsymbol{V}$\tcp*{Randomly sample from a distribution with covariance $\boldsymbol{V}$}
\If{$q^*\notin (q^l,q^u)$}
{
continue\tcp*{Move on if the guess is off the specified limits}
}
\Else
{
$\gamma \sim q^*$\tcp*{Get standard deviation}
\If{$f(\gamma,q^*)$}{
$q^{m+1}\gets q^*$\tcp*{Accept sample based on a condition $f(\gamma,q^*)$ being satisfied}}
\Else{
$q^{m+1}\gets q^m$\tcp*{Reject sample}}
}
$m\gets m+1$\;
\If{$m \% m_0$}{$\boldsymbol{V}\sim \{q^m,q^{m-1},..,q^{m-m_0}\}$\tcp*{Update covariance matrix every $m_0$ iterations}}
}
\end{algorithm}

The adaptive Metropolis algorithm~\cite{haario2001adaptive} explores the parameter space with specified limits ranging from a high of $q^u$ to a low of $q^l$ and starts from a random initial guess of the model parameter $q^{m}$, where $m$ is the iteration number. The initial covariance matrix in the adaptive Metropolis algorithm is constructed using the initial parameter $q^{m=0}$. At each iteration, the steps are
\begin{enumerate}
\item A random parameter sample $q^*$ is generated from the proposal distribution
\item If $q^*$ is not within the specified limits, $q^*\notin (q^l,q^u)$, the iteration is passed without moving to the next steps, and  the previous sample is considered as the new sample, $q^{m+1} = q^{m}$
\item If $q^*$ is within the specified limits, $q^*\in (q^l,q^u)$, a new value of standard deviation associated with $q^*$ is generated using the inverse-gamma distribution
\item $q^*$ is accepted as the new sample $q^{m+1} = q^*$ if a criterion which involves the standard devation is met
\item The covariance matrix is updated if the iteration number is an exact multiple of $m_0$ using the previous $m_0$ model parameters
\end{enumerate}
The simulation is repeated for $n$ iterations and the parameter samples resulting from all these iterations represent the parameter posterior distribution. The construction of the $q^*$ is only based on the current parameter, $q^{m}$, which is the Markov process. The computational time of the MCMC sampling method is proportional to the number of generated samples $n$. To put it more succinctly, the algorithm is elucidated in Algorithm~\ref{mh}.

\section{Results}

The interval of adapting the covariance matrix is 100, which means the matrix is modified every 100 samples. Also, since the initial guess is more often than not way off the desired value, the initial half of the number of samples are burnt-in, which is common practice in MCMC simulations. We start with an initial guess of 1 MSCF/day for all Bayesian/MCMC simulations, which is way off the desired estimated value. In reality, this tests the robustness of the framework, as the initial guess in realistic scenarios is expected to be way off the desired estimated value because we do not know the desired estimated value. 

\subsection{Using the reduced order model obtained through nonoverlapping window approach}

\begin{figure}[htb!]
\begin{subfigure}{.5\textwidth}
    \centering
    \includegraphics[scale=0.35]{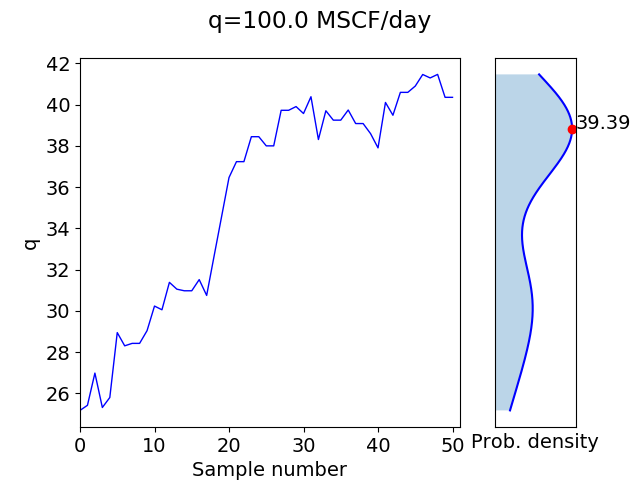}
\end{subfigure}
% \begin{subfigure}{.5\textwidth}
%     \centering
%     \includegraphics[scale=0.35]{pylith_gprs_burn_100.0_200_.png}
% \end{subfigure}
% \begin{subfigure}{.5\textwidth}
%     \centering
%     \includegraphics[scale=0.35]{pylith_gprs_burn_100.0_400_.png}
% \end{subfigure}
\begin{subfigure}{.5\textwidth}
    \centering
    \includegraphics[scale=0.35]{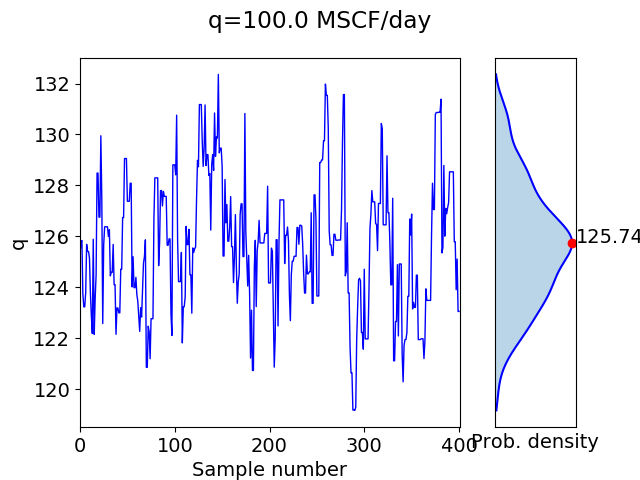}
\end{subfigure}
    \caption{Inversion results for 100 MSCF/day with nonoverlapping window based reconstruction}
    \label{inference1}
\end{figure}

\begin{figure}[htb!]
\begin{subfigure}{.5\textwidth}
    \centering
    \includegraphics[scale=0.35]{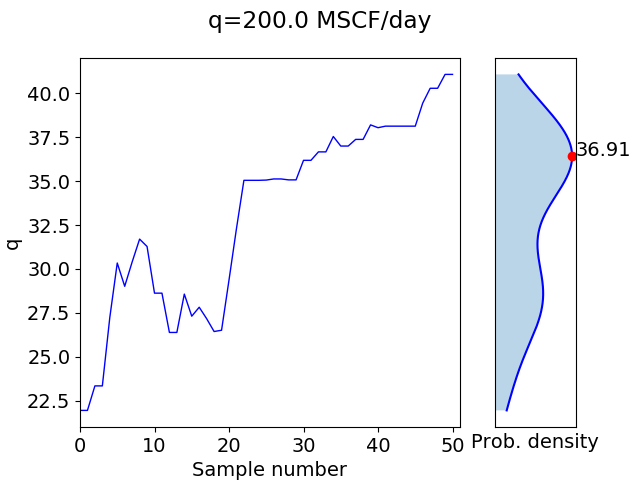}
\end{subfigure}
% \begin{subfigure}{.5\textwidth}
%     \centering
%     \includegraphics[scale=0.35]{pylith_gprs_burn_200.0_200_.png}
% \end{subfigure}
% \begin{subfigure}{.5\textwidth}
%     \centering
%     \includegraphics[scale=0.35]{pylith_gprs_burn_200.0_400_.png}
% \end{subfigure}
\begin{subfigure}{.5\textwidth}
    \centering
    \includegraphics[scale=0.35]{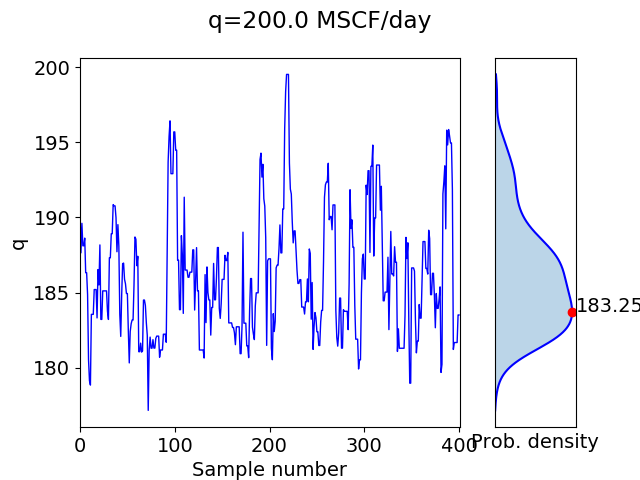}
\end{subfigure}
    \caption{Inversion results for 200 MSCF/day with nonoverlapping window based reconstruction}
    \label{inference2}
\end{figure}

\begin{figure}[htb!]
\begin{subfigure}{.5\textwidth}
    \centering
    \includegraphics[scale=0.35]{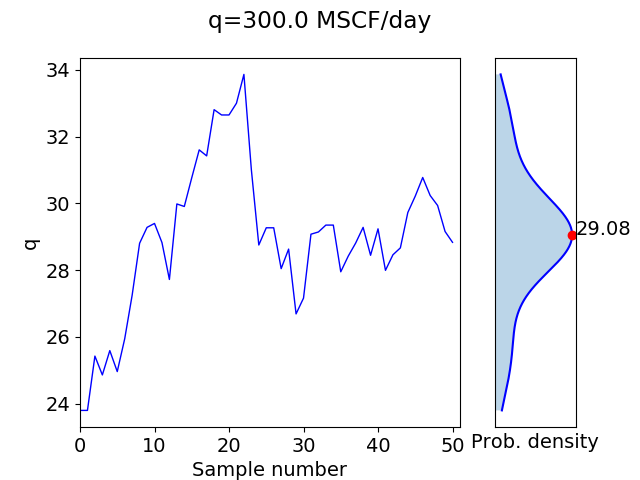}
\end{subfigure}
% \begin{subfigure}{.5\textwidth}
%     \centering
%     \includegraphics[scale=0.35]{pylith_gprs_burn_300.0_200_.png}
% \end{subfigure}
% \begin{subfigure}{.5\textwidth}
%     \centering
%     \includegraphics[scale=0.35]{pylith_gprs_burn_300.0_400_.png}
% \end{subfigure}
\begin{subfigure}{.5\textwidth}
    \centering
    \includegraphics[scale=0.35]{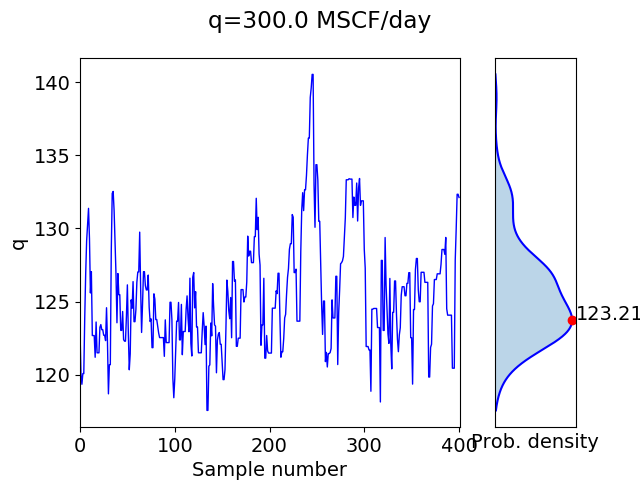}
\end{subfigure}
    \caption{Inversion results for 300 MSCF/day with nonoverlapping window based reconstruction}
    \label{inference3}
\end{figure}

\begin{figure}[htb!]
\begin{subfigure}{.5\textwidth}
    \centering
    \includegraphics[scale=0.35]{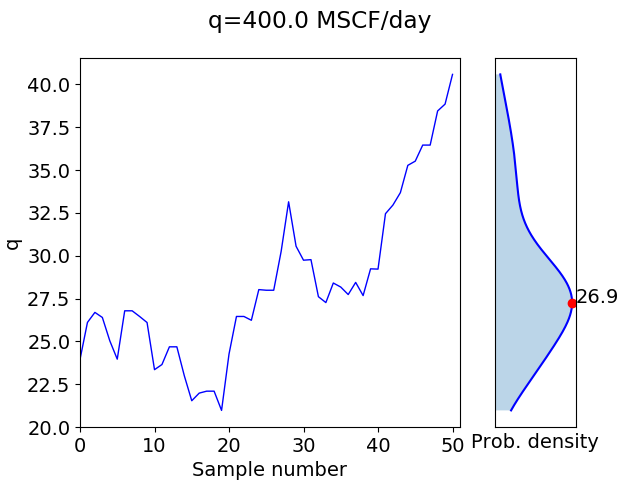}
\end{subfigure}
% \begin{subfigure}{.5\textwidth}
%     \centering
%     \includegraphics[scale=0.35]{pylith_gprs_burn_400.0_200_.png}
% \end{subfigure}
% \begin{subfigure}{.5\textwidth}
%     \centering
%     \includegraphics[scale=0.35]{pylith_gprs_burn_400.0_400_.png}
% \end{subfigure}
\begin{subfigure}{.5\textwidth}
    \centering
    \includegraphics[scale=0.35]{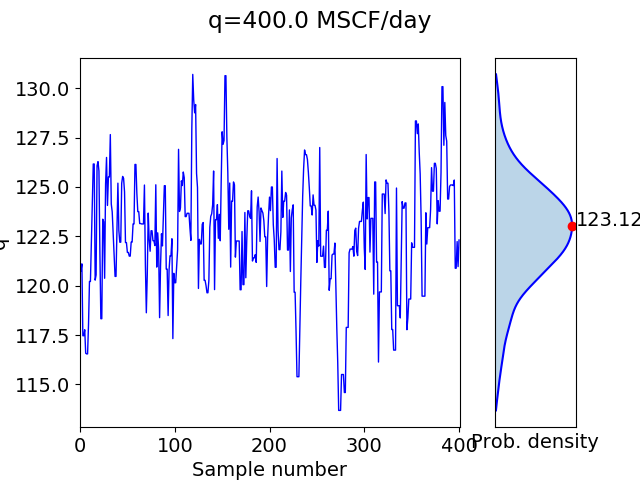}
\end{subfigure}
    \caption{Inversion results for 400 MSCF/day with nonoverlapping window based reconstruction}
    \label{inference4}
\end{figure}

Figs.~\ref{inference1}-~\ref{inference4} are results of the Bayesian/MCMC inference framework for the different injection rates in the spectrum for different number of samples. We observe from the results that the estimation is evidently impacted by how good the reduced order model is in the first place, and we know from Fig.~\ref{reconstruct} that LSTM autoencoders do a lot better when the time series is oscillatory more than any other feature. We also observe that the estimation is not always monotonically converging to the ground truth with increasing number of samples, but is expected to beyond a certain number of samples, which is a function of the value itself, the reduced ordel model, and how well the reduced order model works for the value in and around the ground truth. 

\subsection{Using the reduced order model obtained through sliding window approach}

\begin{figure}[htb!]
\begin{subfigure}{.5\textwidth}
    \centering
    \includegraphics[scale=0.35]{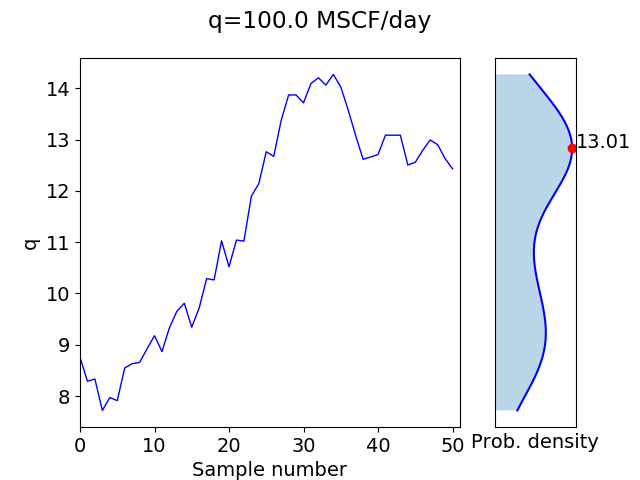}
\end{subfigure}
% \begin{subfigure}{.5\textwidth}
%     \centering
%     \includegraphics[scale=0.35]{pylith_gprs_burn_100.0_200_overlap.png}
% \end{subfigure}
% \begin{subfigure}{.5\textwidth}
%     \centering
%     \includegraphics[scale=0.35]{pylith_gprs_burn_100.0_400_overlap.png}
% \end{subfigure}
\begin{subfigure}{.5\textwidth}
    \centering
    \includegraphics[scale=0.35]{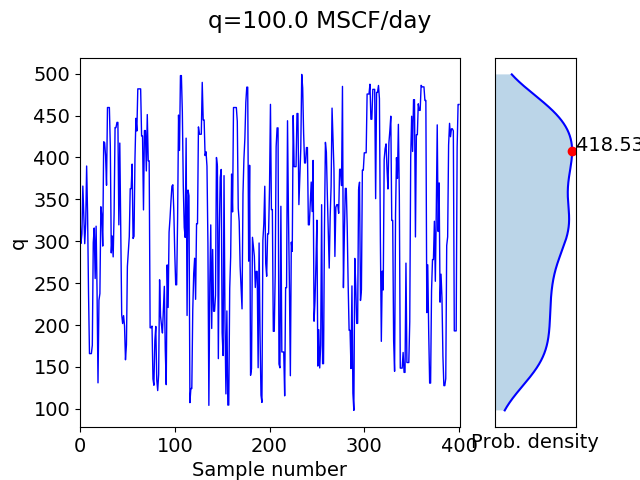}
\end{subfigure}
    \caption{Inversion results for 100 MSCF/day with sliding window based reconstruction}
    \label{inference5}
\end{figure}

\begin{figure}[htb!]
\begin{subfigure}{.5\textwidth}
    \centering
    \includegraphics[scale=0.35]{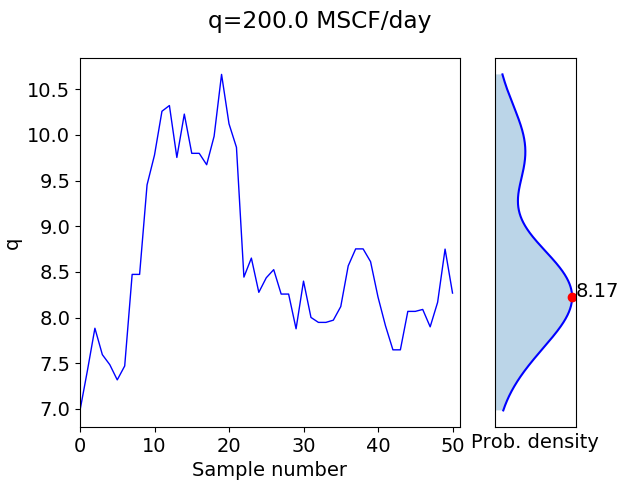}
\end{subfigure}
% \begin{subfigure}{.5\textwidth}
%     \centering
%     \includegraphics[scale=0.35]{pylith_gprs_burn_200.0_200_overlap.png}
% \end{subfigure}
% \begin{subfigure}{.5\textwidth}
%     \centering
%     \includegraphics[scale=0.35]{pylith_gprs_burn_200.0_400_overlap.png}
% \end{subfigure}
\begin{subfigure}{.5\textwidth}
    \centering
    \includegraphics[scale=0.35]{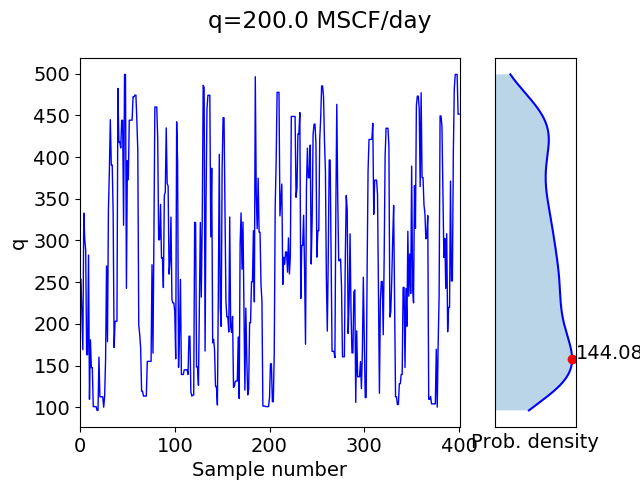}
\end{subfigure}
    \caption{Inversion results for 200 MSCF/day with sliding window based reconstruction}
    \label{inference6}
\end{figure}

\begin{figure}[htb!]
\begin{subfigure}{.5\textwidth}
    \centering
    \includegraphics[scale=0.35]{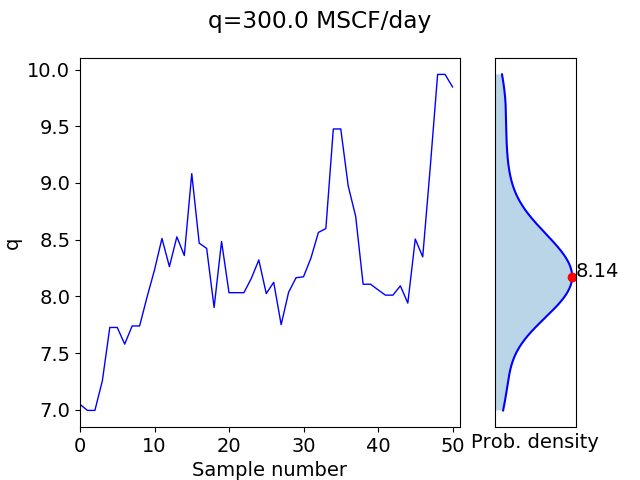}
\end{subfigure}
% \begin{subfigure}{.5\textwidth}
%     \centering
%     \includegraphics[scale=0.35]{pylith_gprs_burn_300.0_200_overlap.png}
% \end{subfigure}
% \begin{subfigure}{.5\textwidth}
%     \centering
%     \includegraphics[scale=0.35]{pylith_gprs_burn_300.0_400_overlap.png}
% \end{subfigure}
\begin{subfigure}{.5\textwidth}
    \centering
    \includegraphics[scale=0.35]{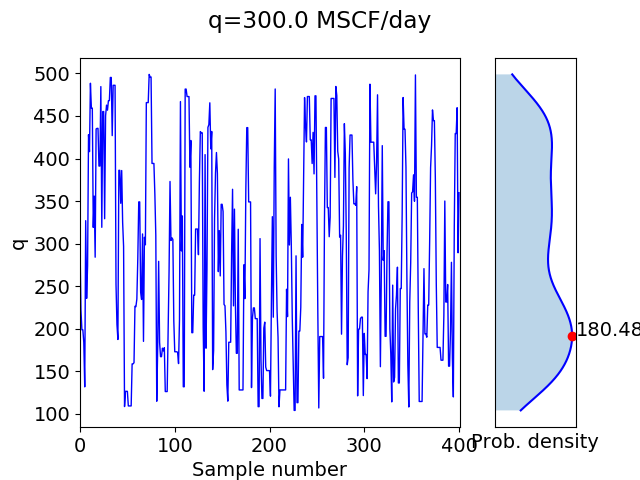}
\end{subfigure}
    \caption{Inversion results for 300 MSCF/day with sliding window based reconstruction}
    \label{inference7}
\end{figure}

\begin{figure}[htb!]
\begin{subfigure}{.5\textwidth}
    \centering
    \includegraphics[scale=0.35]{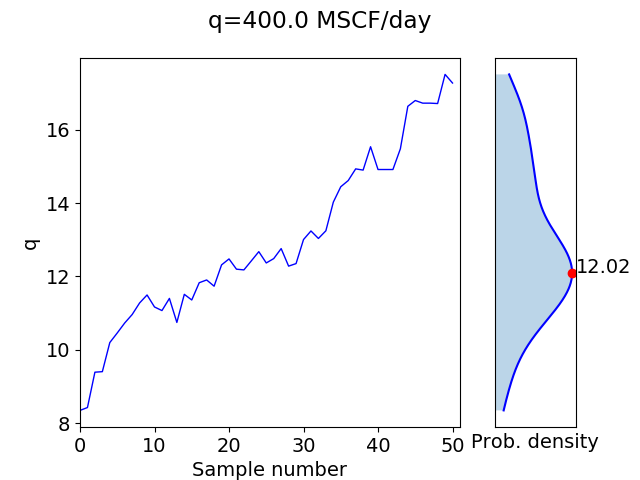}
\end{subfigure}
% \begin{subfigure}{.5\textwidth}
%     \centering
%     \includegraphics[scale=0.35]{pylith_gprs_burn_400.0_200_overlap.png}
% \end{subfigure}
% \begin{subfigure}{.5\textwidth}
%     \centering
%     \includegraphics[scale=0.35]{pylith_gprs_burn_400.0_400_overlap.png}
% \end{subfigure}
\begin{subfigure}{.5\textwidth}
    \centering
    \includegraphics[scale=0.35]{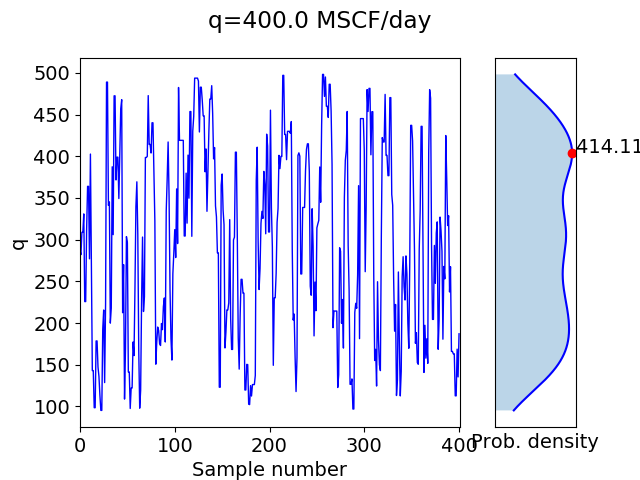}
\end{subfigure}
    \caption{Inversion results for 400 MSCF/day with sliding window based reconstruction}
    \label{inference8}
\end{figure}

Figs.~\ref{inference5}-~\ref{inference8} are results of the Bayesian/MCMC inference framework for the different injection rates in the spectrum for different number of samples. Overall, the estimates with the sliding window based reconstruction is much better than the nonoverlapping window based reconstruction. 

\section{Conclusions and outlook}

To summarise the procedural framework in this work again, the steps we followed were:
\begin{compactitem}
    \item Run high fidelity simulations for a bunch of injection rates
    \item For each injection rate, construct a displacement time series at a chosen grid point on the ground surface
    \item
Add noise to the time series to eventually serve as noisy data for the Bayesian inference framework
\item
Train the LSTM autoencoder with time stamp and injection rate as input and displacement time series (without the noise) as the target
\item
Run the Bayesian inference framework with the LSTM autoencoder based reduced order model
\item
Test the robustness of the framework by comparing the estimates of the injection rate with the ground truth injection rate
\end{compactitem}

We observed that the Bayesian/MCMC performance is squarely a function of how well the reduced order model replicates the high fidelity model, and while this is a good sign as the Bayesian/MCMC piece is robust, it lends to more future work in the realm of doing a good job of model order reduction. In this realm, decompositions play a role in the form of principal component analysis of the time stamps of the solution vector from the high fidelity, and it remains to be seen how to tie in that analysis into a sophisticated forward model using a framework like PyTorch. The author is aware of such frameworks being increasingly developed, and that is the ballpark in terms of future work of developing the software package.

\appendix
\section{The high fidelity forward model}
\label{fmdetails}
\begin{figure}[htb!]
\centering
\includegraphics[trim={0 0 0 0},clip,scale=0.6]{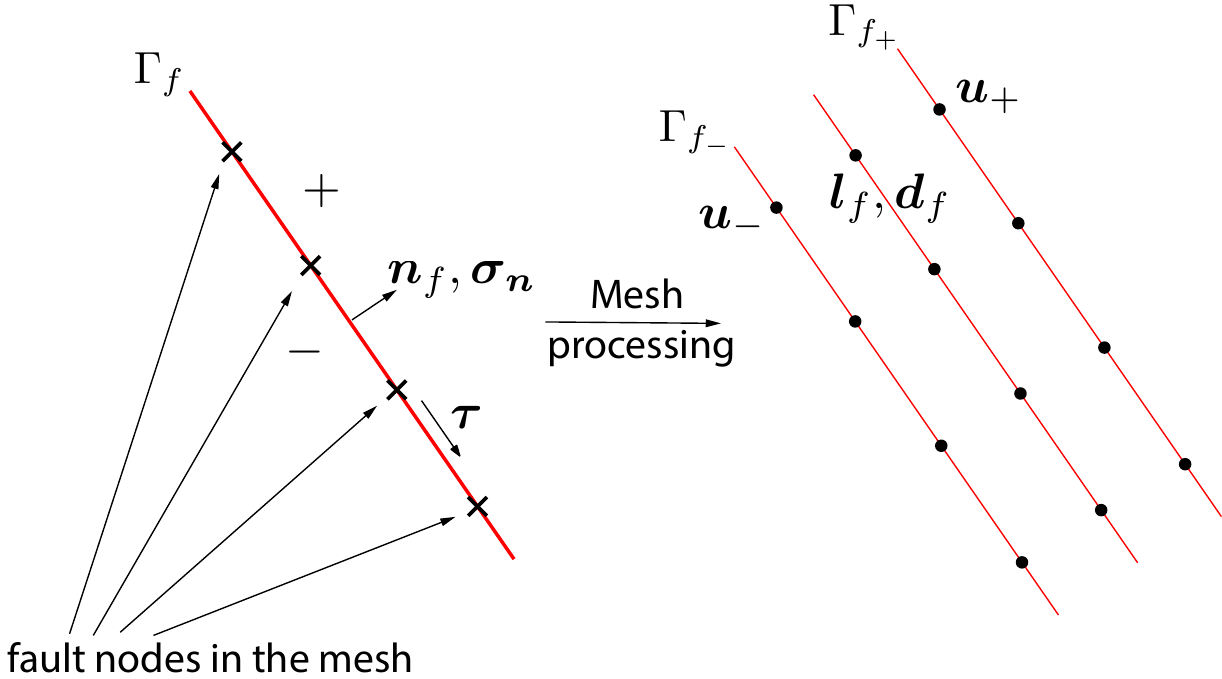}
\caption{A fault surface $\Gamma_f$ is processed to create three surfaces: the positive side surface $\Gamma_{f+}$ containing $\boldsymbol{u}_{+}$, the negative side surface $\Gamma_{f-}$ containing $\boldsymbol{u}_{-}$, and the slip surface containing the fault effective traction vector $\boldsymbol{l}$ and the slip vector $\boldsymbol{d}$}
\label{sketch9}
\end{figure}

The governing PDE for displacement $\mathbf{u}$ is the linear momentum balance given by
\begin{align}
\label{geomechanics}
\nabla\cdot\boldsymbol{\sigma}+\rho_b\mathbf{g}=\mathbf{0}
\end{align}
with the constitutive laws relating poroelastic stress tensor $\boldsymbol{\sigma}$, effective stress tensor $\boldsymbol{\sigma}'$, strain tensor $\boldsymbol{\epsilon}$, volumetric strain $\epsilon=tr(\boldsymbol{\epsilon})$ and pore pressure $p$ given by
\begin{align}
\label{geomechanics0}
\left.\begin{array}{c}
\boldsymbol{\sigma}=\boldsymbol{\sigma}'-b p \mathbf{I}\\
\boldsymbol{\sigma}'=\lambda \epsilon\mathbf{I}+2G\boldsymbol{\epsilon}=\boldsymbol{D}\boldsymbol{\epsilon}
\end{array}\right.
\end{align}
with boundary and initial conditions given by
\begin{align*}
\mathbf{u}=\mathbf{\overline{u}}\,\, \mathrm{on}\,\,\Gamma_D ,\,\,
\dot{\mathbf{u}}=\dot{\mathbf{\overline{u}}}\,\, \mathrm{on}\,\,\Gamma_D,\,\,
\boldsymbol{\sigma}^T\mathbf{n}=\overline{\mathbf{t}}\,\,\mathrm{on}\,\,\Gamma_N\\
\mathbf{u}(\mathbf{x},0)=\mathbf{u}_0(\mathbf{x}),\,\,
\dot{\mathbf{u}}(\mathbf{x},0)=\dot{\mathbf{u}}_0(\mathbf{x})
\end{align*}
As shown in Fig.~\ref{sketch9}, slip on the fault is the displacement of the positive side relative to the negative side:
\begin{align}\label{e:slip}
(\boldsymbol{u}_+ - \boldsymbol{u}_-)-\boldsymbol{d}=\boldsymbol{0} \; \text{on} \; \Gamma_f,
\end{align}
Recognizing that fault tractions are analogous to the boundary tractions, we add in the contributions from integrating the Lagrange multipliers $\boldsymbol{l}\equiv \boldsymbol{\sigma}'\boldsymbol{n}$ over the fault surface in the conventional finite element formulation to get 
\begin{align}
\nonumber
&\int_{\Omega} \nabla\boldsymbol{\eta}: (\boldsymbol{\sigma}'-b p \mathbf{I}) \;d\Omega -\int_{\Omega} \boldsymbol{\eta}\cdot \rho_{b}\boldsymbol{g} \;d\Omega  -\int_{\Gamma_{N}} \boldsymbol{\eta} \cdot  \overline{\boldsymbol{t}}\; d\Gamma\\
\label{blah1}
&+\int_{\Gamma_{f_+}}\boldsymbol{\eta} \cdot (\boldsymbol{l}-bp_{+}\boldsymbol{n})\; d\Gamma  -\int_{\Gamma_{f_-}}\boldsymbol{\eta} \cdot (\boldsymbol{l}-bp_{-}\boldsymbol{n})\; d\Gamma= 0    
\end{align}
We use the Mohr-Coulomb theory~\cite{JaeJ1979} to define the stability criterion for the fault and define a fault pressure $p_f=\frac{p_-+p_+}{2}$. The shear stress and frictional stresses on the fault are
\begin{align*}
\tau&=|\boldsymbol{l}-\sigma'_n\boldsymbol{n}|\equiv |\boldsymbol{l}-(\boldsymbol{l}\cdot\boldsymbol{n})\boldsymbol{n}|
\\
\tau_f &= 
\begin{cases} 
\tau_c - \mu_f \boldsymbol{l}\cdot \mathbf{n},  &\boldsymbol{l}\cdot \mathbf{n} < 0, \\
\tau_c,  &\boldsymbol{l}\cdot \mathbf{n}\ge 0
\end{cases}
\end{align*}
where $\tau_c$ is the cohesive strength of the fault, $\mu_f$ is the coefficient of friction which evolves as
\begin{align}
\label{e:slipweak}
\mu_f = 
\left\{
\begin{array}{c}
\mu_s-(\mu_s-\mu_d) \frac{|\boldsymbol{d}|}{d_c},\quad |\boldsymbol{d}| \le d_c,\\
\mu_d,\quad |\boldsymbol{d}|>d_c
\end{array}\right.
\end{align}
where $d_c$ is a critical slip distance.
\begin{algorithm}[htb!]
\caption{Time-marching in poroelastostatics}
\label{poroelastostatics}
$n\gets 0$; $t\gets 0$\;
$\boldsymbol{d}_h^{0}\gets \boldsymbol{0}$\tcp*{Initialize fault slip at virgin state}
$\boldsymbol{u}_h^{0}\gets \boldsymbol{u}_{h}^{Prestep}$\tcp*{Initial condition based on a elastic prestep solve}
\While{$t<T$}{\tcc{time marching till final time}
\While{Not converged}{\tcc{Staggered solution algorithm loop}
Solve flow problem for pressures\;
$\boldsymbol{d}_h^{n+1}\gets \boldsymbol{d}_h^n$\tcp*{Initialize fault slip for next time step}
Solve system of aligns using GMRES\tcp*{Krylov subspace solver}
$\boldsymbol{L}\gets \boldsymbol{L}+d\boldsymbol{L}$\tcp*{Update lagrange multipliers}
$\boldsymbol{U}\gets \boldsymbol{U}+d\boldsymbol{U}$\tcp*{Update displacements}
\For{Loop over lagrange nodes}{
$\tau\gets |\boldsymbol{l}_h^{n+1}-(\boldsymbol{l}_h^{n+1}\cdot\boldsymbol{n})\boldsymbol{n}|$\tcp*{Obtain shear stress on fault}
$\tau_f\gets \tau_f\vert_{\boldsymbol{d}_h^{n}}$\tcp*{Compute fault friction based on the slip value at previous time step}
\While{$\tau>\tau_f$}{\tcc{Satisfy fault constitutive law}
$\boldsymbol{U}_+\gets \boldsymbol{U}_+-\boldsymbol{K}_{++}^{-1}\frac{\tau-\tau_f}{\tau}\boldsymbol{L}$\;
$\boldsymbol{U}_-\gets \boldsymbol{U}_-+\boldsymbol{K}_{--}^{-1}\frac{\tau-\tau_f}{\tau}\boldsymbol{L}$\;
$\boldsymbol{d}_h^{n+1}\gets \boldsymbol{u}_{h_+}^{n+1}-\boldsymbol{u}_{h_-}^{n+1}$\tcp*{Update fault slip}
$\tau_f\gets \tau_f\vert_{\boldsymbol{d}_h^{n+1}}$\tcp*{Update fault friction}
}
$\boldsymbol{d}_h^{n+1}\gets \boldsymbol{u}_{h_+}^{n+1}-\boldsymbol{u}_{h_-}^{n+1}$\tcp*{Update fault slip}
}
}
$n\gets n+1$\;
$t\gets t+\Delta t$\;
}
\end{algorithm}
The fields are approximated as follows:
\begin{align*}
\boldsymbol{u} \approx \boldsymbol{u}_h=\sum_{b=1}^{n_{\text{node}}}\eta_b\boldsymbol{U}_b,\quad\boldsymbol{l} \approx \boldsymbol{l}_h=\sum_{b=1}^{n_{f,\text{node}}}\eta_{b}\boldsymbol{L}_{b},\quad
\boldsymbol{d} \approx \boldsymbol{d}_h=\sum_{b=1}^{n_{f,\text{node}}}\eta_{b}\boldsymbol{D}_{b},
\end{align*}
where $n_{\text{node}}$ is the total number of nodes and $n_{f,\text{node}}$ is the number of Lagrange nodes. After substitution of the finite element approximations into the weak form of the problem, we obtain the fully discrete aligns in residual form for all nodes $a$ and lagrange nodes $\bar{a}$:
\begin{align}
\nonumber
\boldsymbol{R}_{u,a}^{Stat} &= \int_{\Omega} \boldsymbol{B}_a^T (\boldsymbol{\sigma}_h'^{n+1}-bp_h^{n+1}\boldsymbol{1}) d\Omega
-\int_{\Omega} \boldsymbol{\eta}_a^T\rho_{b,h}^{n+1}\boldsymbol{g} d\Omega -\int_{\Gamma_{N}} \boldsymbol{\eta}_a^T\overline{\boldsymbol{t}} d\Gamma\\
\label{res1_stat}
&+\int_{\Gamma_{f_+}}\boldsymbol{\eta}_{a}^T(\boldsymbol{l}_h^{n+1}-bp_{f,h}^{n+1}\boldsymbol{n}) d\Gamma-\int_{\Gamma_{f_-}}\boldsymbol{\eta}_{a}^T(\boldsymbol{l}_h^{n+1}-bp_{f,h}^{n+1}\boldsymbol{n}) d\Gamma=\boldsymbol{0}\\ 
\label{res2_stat}
\boldsymbol{R}_{l,\bar{a}}^{Stat} &= \int_{\Gamma_{f_+}}\boldsymbol{\eta}_{\bar{a}}^T\boldsymbol{u}_{h_+}^{n+1}d\Gamma - \int_{\Gamma_{f_-}}\boldsymbol{\eta}_{\bar{a}}^T\boldsymbol{u}_{h_-}^{n+1}\;d\Gamma - \int_{\Gamma_f}\boldsymbol{\eta}_{\bar{a}}^T\boldsymbol{d}_h^{n+1} d\Gamma=\boldsymbol{0}
\end{align}
We find the system Jacobian matrix by isolating the term for the increments in displacements and Lagrange multipliers at time step $n+1$. The system of linear aligns is:
\begin{align}\label{e:linmech1_stat}
  \left[\begin{array}{ccc:c}
    \boldsymbol{K}_{rr} & \boldsymbol{K}_{r+} & \boldsymbol{K}_{r-}   & \boldsymbol{0}   \\
    \boldsymbol{K}_{+r} & \boldsymbol{K}_{++} & \boldsymbol{0}   & \boldsymbol{C}_{+}^T \\
    \boldsymbol{K}_{-r} & \boldsymbol{0} & \boldsymbol{K}_{--}   & -\boldsymbol{C}_{-}^T \\ \hdashline
    \boldsymbol{0} & \boldsymbol{C}_{+} & -\boldsymbol{C}_{-}   & \boldsymbol{0}
  \end{array}\right]
  \begin{bmatrix}
    d\boldsymbol{U}_{r} \\
    d\boldsymbol{U}_{+} \\
    d\boldsymbol{U}_{-} \\
    \hline
    d\boldsymbol{L}  
  \end{bmatrix} = -  
  \begin{bmatrix}
    \boldsymbol{R}_{u,r}^{Stat} \\
    \boldsymbol{R}_{u,+}^{Stat} \\
    \boldsymbol{R}_{u,-}^{Stat} \\
    \hline
    \boldsymbol{R}_{l}^{Stat}
  \end{bmatrix}
\end{align}
where the top row corresponds to displacement nodes excluding the fault positive and negative side nodes. In many quasi-static simulations it is convenient to compute a static problem with elastic deformation prior to computing
a transient response. The heavy lifting for the time marching is done at the Krylov solver~\cite{krylov} stage to solve the system of Eqs.~\eqref{e:linmech1_stat} as shown in Algorithm~\ref{poroelastostatics}. 

\bibliographystyle{elsarticle-num-names} 
\bibliography{bib}

\end{document}